\documentclass[letterpaper, 10 pt, conference]{ieeeconf}  % Comment this line out if you need a4paper

\IEEEoverridecommandlockouts                              % This command is only needed if 
\usepackage{graphicx}
\usepackage[normalem]{ulem}
\usepackage{amsmath,amssymb}

\usepackage{upgreek}

\usepackage{xcolor}
\usepackage{cite}

\usepackage{tabularx}
\usepackage{booktabs}
\usepackage{rotating}
\usepackage{array}
\usepackage{bm}

\usepackage{pbalance}

\usepackage{listings}
\title{\LARGE \bf
PHD-Hand: A Printable Humanoid Dexterous Hand with Linkage-Driven 3-DOA Digits
}

\author{Neel Adwani\,\textsuperscript{\textit{a}}, Muhaiminul Islam Akash\,\textsuperscript{\textit{a}}, Saffia Ouali\,\textsuperscript{\textit{a}}, Ethan Liu\,\textsuperscript{\textit{b}}, Cong Wang\,\textsuperscript{\textit{a}}% <-this % stops a space
\thanks{
\textsuperscript{\textit{a}} Electrical and Computer Engineering, New Jersey Institute of Technology (NJIT) {\tt\small \{na657, ma2693, so22, wangcong\}}@njit.edu
} 
\thanks{
\textsuperscript{\textit{b}} Union County Magnet High School, USA, ethanliu030@gmail.com
}%
}

\begin{document}

\maketitle
\thispagestyle{empty}
\pagestyle{empty}

%%%%%%%%%%%%%%%%%%%%%%%%%%%%%%%%%%%%%%%%%%%%%%%%%%%%%%%%%%%%%%%%%%%%%%%%%%%%%%%%
\begin{abstract}
Currently, dexterous humanoid robot hands are widely considered the final hardware bottleneck that hampers generalist robots from physically assisting people at home and work. 
% As conventional tendon-driven designs suffer from high cost, demanding maintenance, poor reliability, and short durability, alternative linkage-driven designs have gained attention increasingly. 
As conventional tendon-driven designs suffer from high cost, demanding maintenance, poor reliability, and short durability, alternative linkage-driven designs have increasingly gained attention. 
% Despite their potentials to avoid the weaknesses of tendon-driven hands, integrating multiple degrees of actuation (DOA) in a single digit with linkage drivetrains is a challenge for design and fabrication. 
Despite their potential to avoid the weaknesses of tendon-driven hands, integrating multiple degrees of actuation (DOA) in a single digit with linkage drivetrains is a challenge for design and fabrication.
Only a few studies have managed to present functional humanoid hands that are completely linkage-driven and have at least three DOAs per digit without compromising actuator placement and other design preferences.  Our work introduces a new linkage-driven hand featuring a unique flexure hinge.  Together with elastic drive rods made of spring steel, the linkage drivetrains are significantly simplified.  Unlike other linkage-driven designs of similar dexterity, our design does not require many miniature moving parts.  As a result, the hand can be fabricated easily with a budget 3D printer.  This not only significantly lowers the barrier to entry for using and developing dexterous humanoid robot hand functions, but also allows the hands to be sized much closer to regular human hands.  The design offers desirable kinematic advantages as well.  A prototype is evaluated for grasping and object manipulation.  In addition, this paper presents our latest findings on kinematics and control of the hand.

\end{abstract}


%%%%%%%%%%%%%%%%%%%%%%%%%%%%%%%%%%%%%%%%%%%%%%%%%%%%%%%%%%%%%%%%%%%%%%%%%%%%%%%%
\section{Introduction}
\label{sec introduction}
This paper discusses the design of linkage-driven dexterous humanoid robot hands. 
With the recent breakthroughs in perception and reasoning powered by artificial intelligence~(AI), replicating the dexterity of human hands is increasingly considered the last physical bottleneck that holds back the robotics industry.
Having humanoid dexterous hands not only allows robots to seamlessly handle objects in our daily lives, but also enables robot learning of hand skills directly from native human demonstration.
Other than resembling the appearance of human palm and digits\footnote{
In human anatomy, the thumb is not a finger - i.e., a regular human hand has five digits, including one thumb and four fingers.},
%a humanoid robot hand ideally should have at least three actuated degrees of freedom (more precisely, degrees of actuation or DOA) per digit.
a humanoid robot hand ideally should have at least three \uline{degrees of actuation (DOA)}\footnote{
Because of the complex kinematics of dexterous robot hands, the term ``degrees of freedom (DOF)'' has been used to indicate quite different meanings in the literature.
Instead, we have found ``degrees of actuation (DOA)'' to be a more precise term.
}
per digit.
The ``\uline{root joint}'' of a digit should have two DOAs for tilting, while the ``\uline{upper two joints}''\footnote{
For a finger, the root joint is an MCP joint, and the upper two joints include a PIP joint and a DIP joint.
For the thumb, the root joint is a CMC joint, and the upper two joints include an MCP joint and an IP joint.
To keep the narrative concise and friendly to a broader audience, instead of the anatomical terminology, we use the terms ``root joint'' and ``upper two joints'' in this paper.
}
almost always flex/extend together and share a single DOA\footnote{
The thumb actually moves differently from the fingers, though the difference is not that significant.
}.

A straightforward way to build such a digit is to place the actuators inside or on the digits to drive the joints ``locally'' with a structure similar to serial robot manipulators (e.g.,~\cite{shaw2023leap,hernando2023mechatronic,romero2024eyesight,mouri2002anthropomorphic}).
%(e.g.,~\cite{shaw2023leap}\cite{hernando2023mechatronic}\cite{romero2024eyesight}\cite{mouri2002anthropomorphic}).
This results in a dilemma of the digit either having weak actuation or being oversize and overweight.
It is therefore desirable to have the actuators placed off the digits, which requires additional drivetrain(s) to deliver actuation to the upper two joints.
The most common solution is to use tendon drives.
The design is largely inspired by human anatomy and has been the mainstream choice of dexterous robot hands~\cite{piazza2019century}.
% However, without self-healing, the complex tendon systems required by the dexterity wear out and loosen up quickly, which leads to demanding maintenance and short life span~\cite{teslahand}.
However, without self-healing, the complex tendon systems required for dexterity wear out and loosen up quickly, which leads to demanding maintenance and short life span~\cite{teslahand}.
Performance of tendon-driven hands is often jeopardized by cable fraying, knot slippage, and tension drift~\cite{bhatt}\cite{jeong}.
In particular, cable rupture is the top failure mode~\cite{jeong}.
In addition, tendon-driven bidirectional actuation requires either two actuators per DOA (leads to sizing problems) or closed-loop tendon routing (prone to loosening and slacking)~\cite{iitsofthand}.
The absence of bidirectional actuation excludes the hand from performing certain skills.
%such as flicking and thumbing. 

Compared to tendon-drives, rigid drivetrains do not have weaknesses mentioned above.
They also provide better precision and can bear higher load~\cite{wangandli}.
Some designs use geared drivetrains to transmit actuation from actuators placed off the digit to the upper two joints.
\cite{ueda2005development} shows a representative design in this category.
Each of its digits has a 2-DOA root joint and 1-DOA coupled upper two joints.
Actuation is transmitted to the upper two joints using a bevel gear drive coaxially located with the drive shafts of the root joint. %(Fig.~\ref{fig MCP-prior-work}-b).
Other than the fabrication hassle and inevitable backlash, complexity of the design results in a hand size significantly larger than human hands.

\begin{figure}[thpb]
      \centering
      \includegraphics[width=3.4in]{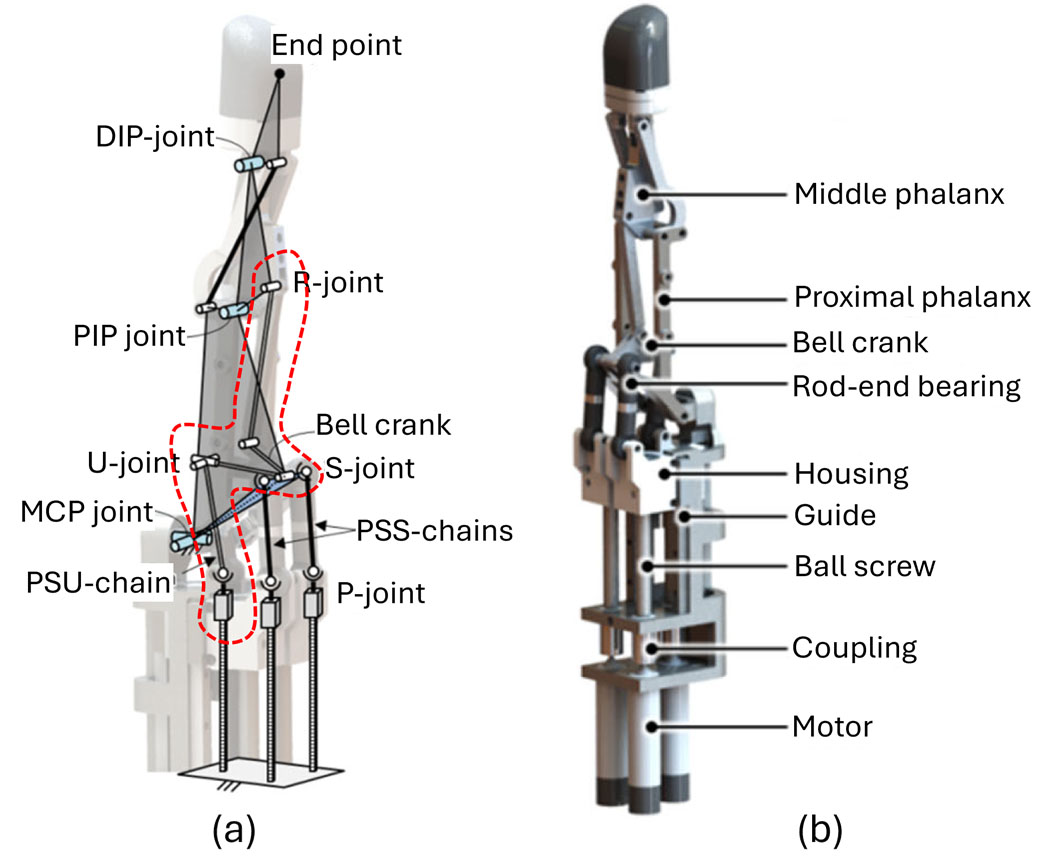}
      \caption{A representative existing design of linkage-driven 3-DOA digit~\cite{kim2021integrated}, a - schematic, b - parts and components}
      \label{fig MCP prior work}
\end{figure}

Alternatively, linkage-driven designs have the potential to avoid the weaknesses of tendon and gear-based designs~\cite{kashef2020robotic}.
However, integrating three DOAs with linkage drivetrains in one digit (while keeping the actuators off the digit) poses a great challenge for design and fabrication.
%especially in terms of transmitting actuation to the upper two joints through the root joint.
There are only a few presentations~\cite{kim2021integrated,wen2025dexterous,wen2025gr,liu2022three,zaw2025design}, which share similar designs and have only emerged over recent years.
Among them, the design presented in \cite{kim2021integrated} is both pioneering and most ready for practical use.
It adopts a novel 5-joint linkage drivetrain to deliver actuation to the upper two joints through the root joint (Fig.~\ref{fig MCP prior work}-a, the portion circled in red).
As impressive as the prototype and demonstration are, the nontrivial number of miniature parts (Fig.~\ref{fig MCP prior work}-b) require serious fabrication capability and make it challenging to balance between durability, size, and cost.
The prototype in~\cite{kim2021integrated} has a hand length around 218 mm which is 13\% larger than average male human hands.
The design is replicated in \cite{wen2025dexterous} and improved in \cite{wen2025gr} with hand lengths of 255 mm and 219 mm respectively.
In addition, the kinematics of the root joint and upper two joints is partially coupled, which complicates the control.
More details of these designs will be examined later in Section~\ref{sec design} with a summarizing comparison to our work.
Some other designs work around the challenge by strategically reducing the number of DOAs based on human hand synergies~-~e.g.,~\cite{cerulo2017teleoperation,yoon2017underactuated,liu2017development,hong2018kulex}.
Despite being able to realize a decent variety of hand functions, the strategy does not provide the potential of advanced dexterity that the robotics community is currently wishing for.

Our work aims at developing a dexterous humanoid robot hand that 
\begin{itemize}
    \item Has five digits with three DOAs per digit (two for the root joint, one for the coupled upper two joints); 
    \item Has each DOA bidirectionally actuated with a decent strength, good backdrivability, and minimal backlash;
    \item Can be fabricated at a size very close to human hands with minimal costs using a budget 3D printer; 
    %and generic hardware and servos commonly available on the market; 
    \item Has long-lasting durability and requires minimal maintenance.
\end{itemize} 
As explained earlier, these requirements exclude conventional designs that use tendons, gears (other than in the actuators), or place actuators inside/on the digits, and leaves a full linkage-driven design the most feasible option.
Thanks to a unique design of the root joint, our hand has much fewer parts than existing designs that have a similar dexterity, and can be sized closer to human hands.
In addition, our design also completely decouples the kinematics of the root joint and the upper two joints, which simplifies modeling, estimation, and control.

This paper continues from our earlier publication~\cite{adwani2026dexterous}, which introduces the flexure design of the root joint and presents a preliminary prototype of three digits, together with strength test results.
Our prior work suffers from a series of problems, including mechanical interference, non-humanoid topology, and fabrication complications.
%including limited motion range caused by mechanical interference, non-humanoid topology caused by suboptimal placement of the servos, and complications in the fabrication.
The new design in this paper resolves those problems and demonstrates legitimate humanoid dexterity.
In addition, this paper presents our latest findings on kinematics and control of the hand.

% Custom column type for wrapped, top-aligned text
\newcolumntype{L}[1]{>{\raggedright\arraybackslash}p{#1}}
\begin{table*}[ht]
\centering
\scriptsize
\renewcommand{\arraystretch}{1.3}
\caption{Comparison between the proposed design and its peers}
\label{tab:hand_comparison}
\begin{tabular}{|L{2.5cm}|L{2.5cm}|p{2.5cm}|L{2.5cm}|L{2.5cm}|L{2.5cm}|}
\hline
\textbf{Design} & This work & \cite{kim2021integrated} \newline \cite{wen2025dexterous} (replicated) \newline \cite{wen2025gr} (improved) & \cite{liu2022three} & \cite{zaw2025design} & L20 by Linkerbot \newline (commercial product) \\
%https://linkaihand.com/products
\hline
\textbf{Release year} & 2026 & 2021, 2025, 2025 & 2022 & 2024 & after 2023 \\
\hline
\textbf{Digits} & 1 thumb + 4 fingers & 1 thumb + 4 fingers & 1 thumb + 2 fingers & Single digit only & 1 thumb + 4 fingers \\
\hline
\textbf{DOAs / digit} & 3 & 3~\cite{kim2021integrated}\cite{wen2025dexterous} \newline 3 each finger and 4 on the thumb~\cite{wen2025gr} & 3 each finger and 4 on the thumb & 3 & 3 each finger, 4 on the thumb \\
\hline
\textbf{Major parts / digit
(including digit body and drivetrain but not actuators and sensors)} & 13 if the phalanges and digit joints are made in one piece; 18 if made separate parts & 27+ & 27+ & (No physical demonstration yet) & (unknown) \\
\hline
\textbf{Fabrication requirement} & DIY with a budget 3D printer & industrial & industrial & (No physical demonstration yet) & industrial \\
\hline
\textbf{Kinematics of root joint and upper two joints} & Completely decoupled & Partially coupled & Partially coupled & Partially coupled & (unknown) \\
\hline
\textbf{Backdrivability} & Yes & No & No & (unknown) & (unknown). \\
\hline
\textbf{Physical scale} & Whole hand close to regular human (200 mm hand length). & Whole hand larger than regular human (218, 255, 219 mm hand length respectively). & Only finger length (about 80 mm) close to regular human. Whole hand is visibly larger. & (No physical demonstration yet) & Whole hand significantly larger than human (around 225 mm hand length). \\
\hline
\end{tabular}
\end{table*}

\section{Design and Fabrication}
\label{sec design}
\begin{figure}[ht]
      \centering
      \includegraphics[width=3.3in]{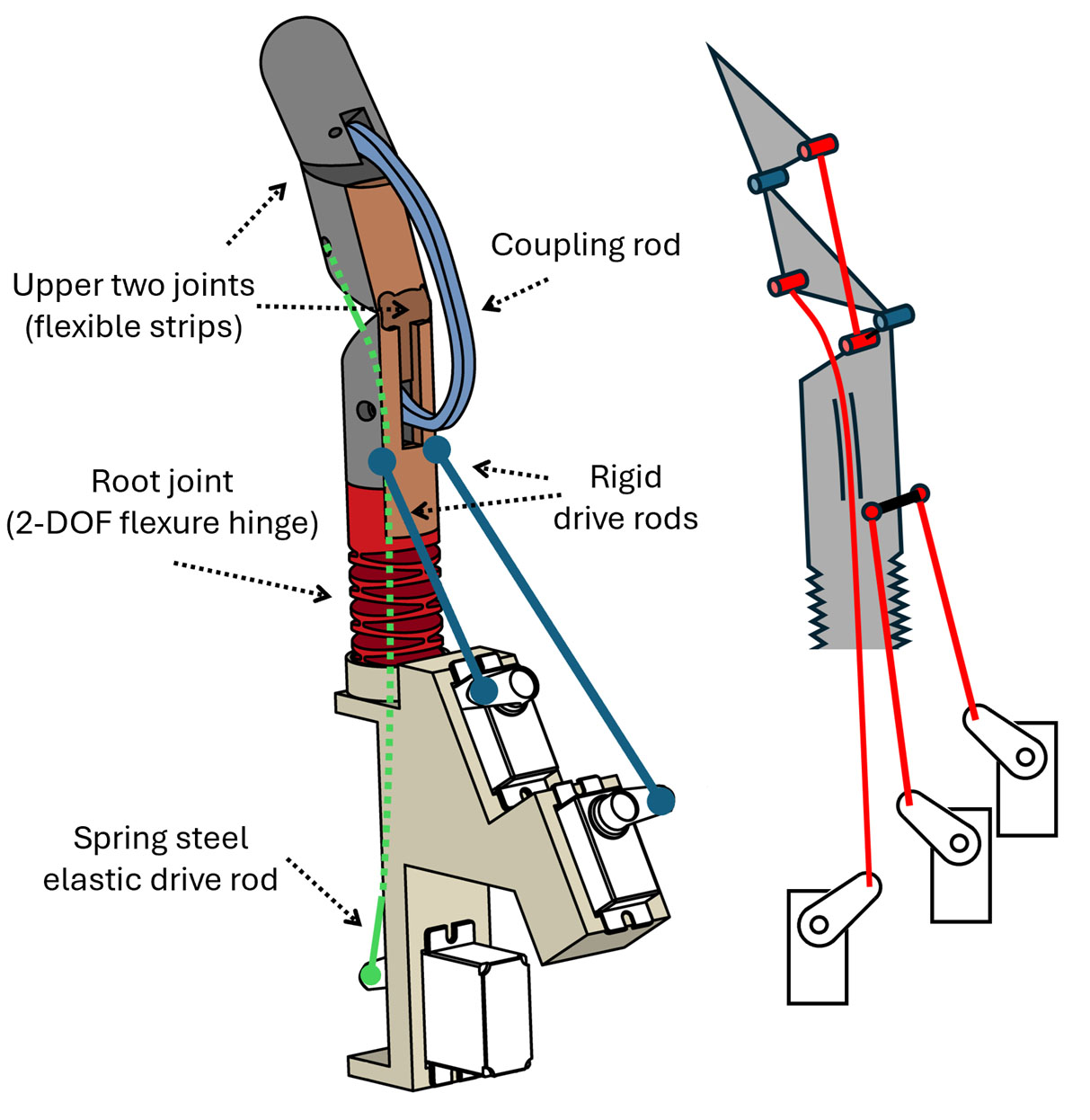}
      \caption{The proposed design of a digit}
      \vspace{-0.5cm}
      \label{fig design}
\end{figure}

Figure~\ref{fig design} illustrates our proposed design.
The root joint is the key element.
We present a unique design in the form of a flexure hinge.
The design is inspired by stress-relieving structures commonly used to protect cable connectors.
As a single-piece anisotropic flexible structure, it provides two DOFs of tilting motion while restraining axial rotation.
It gives a wide motion range to satisfy the need of dexterity, yet does not bend sharply.
This is an important feature that allows an elastic drive rod made of spring steel to go through its center and smoothly deliver bidirectional actuation to the upper two joints without the need of complicated mechanisms and many moving parts (as opposed to the designs in~\cite{kim2021integrated,wen2025dexterous,wen2025gr,liu2022three,zaw2025design}).
In addition, bending of the flexure hinge does not force/require the elastic rod to move axially.
This completely decouples the drivetrain kinematics of the root joint and the upper two joints, which greatly simplifies modeling, estimation, and control.
%Full kinematic decoupling between the root joint and the two distal joints, the root joint is actuated by two servos, while the distal joints are driven independently by a third;
%\item Withstand high structural loads at the finger's base joint, which disqualifies mechanically complex solutions.
In terms of actuation, the root joint is driven by two RC servos that connect to the root phalanx of the digit through two rigid drive rods and four ball joints.
The design evolves from the classic 2-DOF planar 5-bar mechanisms.
% It allowed us to reduce the horizontal footprint, making it easier to move multiple fingers closer
%Together with the horns of the servos, they form a 5-bar mechanism.
Note that the axial rotation of the drive rods is not constrained because of the ball joints, though it does not affect the function of the drivetrain.
The upper two joints of the digit are coupled by a cross 4-bar mechanism, which is a commonly used design. 
Reported coupling ratios of the upper joints of human digits fall between 0.75 and 0.8~\cite{roda2022studying}.
We find a ratio of 0.77 optimal for our design.
The cross 4-bar mechanism is actuated by the elastic drive rod that passes through the root joint.
As of now, we have decided to make the coupling rod go from outside of the middle phalanx.
Other than minimizing mechanical interference and providing a bigger motion range, the option allows easy shaping of the coupling rod to give it certain flexibility, which helps realize the passive compliance desired in hand-object interaction~\cite{ma2014underactuated}\cite{kuo2010contribution}.

\begin{figure}[ht]
      \centering
      \includegraphics[width=2.9in]{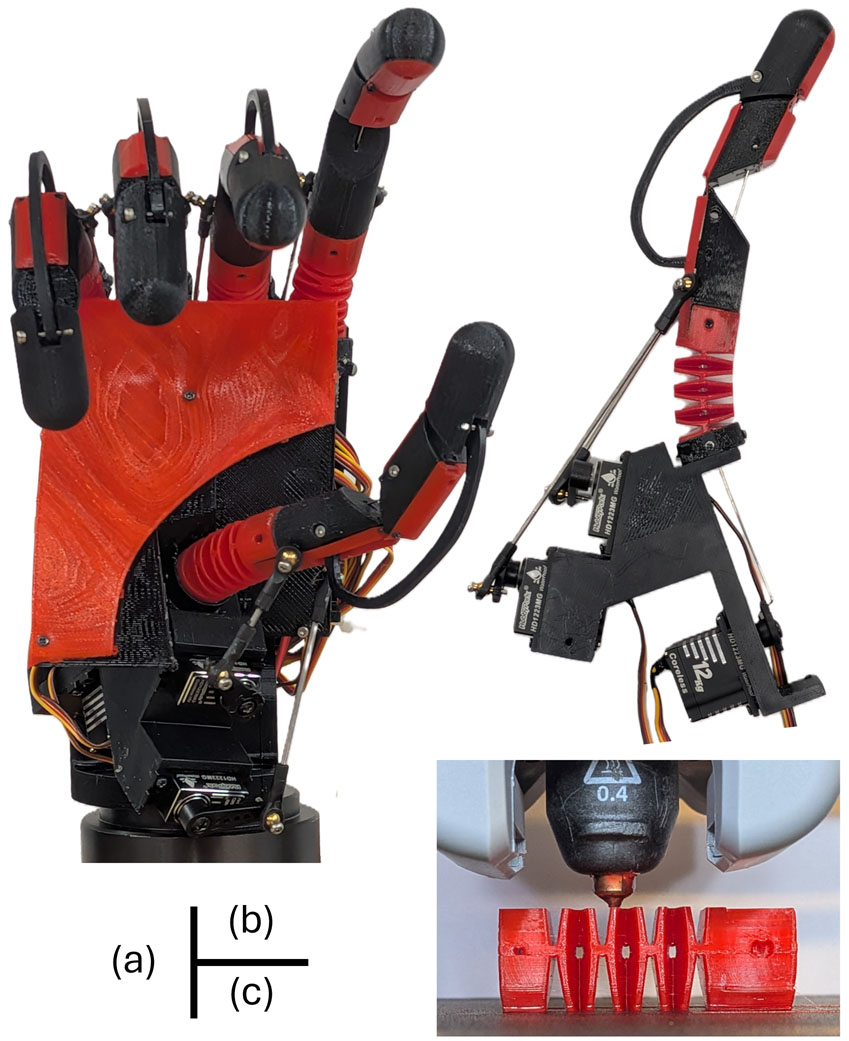}
      \caption{Prototype - (a) whole hand, (b) modular digit, (c)~3D~printing of the 2-DOF flexure hinge}
      \vspace{-0.3cm}
      \label{fig prototype}
\end{figure}

Figure~\ref{fig prototype}-a shows a prototype.
The digits are modular (Fig.~\ref{fig prototype}-b), allowing easy reconfiguration of the hand topology and repair.
Thanks to the unique design, the hand can be fabricated using a budget FFF 3D printer, generic hardware (ball joints, rods, pins, and fasteners), and generic mini-scale RC servos commonly available on the market of crafting hobbies.
Because of its nontrivial shape, 3D printing of the root joint can be tricky but can still be done reliably with carefully tuned settings.
We recommend using TPU 95A filament, a 0.4 mm nozzle, a nozzle temperature of 215 °C, a bed temperature of 55 °C with an enclosure, a printing speed of 100 mm/s, a layer height of 0.15 mm, and 15\% infill, with the print placed horizontally without any supports~(Fig.~\ref{fig prototype}-c).
The upper two joints are made as flexible strips 3D printed with TPU.
The phalanges can be either 3D printed with PETG for higher rigidity or with TPU so that the whole digit body can be made in one single piece to 
%reduce the total number of parts from 18 to 13 per digit and 
further simplify assembly.
The elastic drive rod is cut from 304 stainless spring steel wires of 0.6 mm thickness.
It is flexible enough to bend with the root joint without causing significant resistance, while is strong enough to avoid buckling when pushed.

Our literature survey has found four existing designs that are fully linkage-driven, have at least three DOAs per digit, and with all actuators placed off the digits.
Three of them are presented in research papers while the other one is a commercial product.
As detailed in Table~\ref{tab:hand_comparison}, the parts of our design are not only significantly fewer but also can be affordably made without industrial-level fabrication.
The advantage greatly lowers the barrier to entry for the use and development of dexterous robot hand functions. 
The concise design also allows the hand to be made at a more compact size closer to regular humans.
Instead of the custom linear actuators with leadscrews used in other designs, our design uses generic RC servos.
Despite giving a raised back-palm, the choice not only simplifies fabrication but also provides a good backdrivability.

\section{Kinematics Analysis}
\label{sec kinematics}
Thanks to our unique design, kinematics of the root joint and the upper two joints is completely decoupled.
That is, when the root joint tilts, the servo that actuates the upper two joints does not need to move coordinately to preserve their flexion, and vice versa. 
Kinematics of the cross 4-bar mechanism that drives the upper two joints is simple and well documented in textbooks.
This section focuses on discussing the kinematics of the root joint.
With its parallel drivetrain, the 2-DOA root joint system is a closed-chain mechanism, which means its kinematics cannot be derived easily using a DH-parameter model.
Below, we first present a particular solution to derive its inverse kinematics.

\begin{figure}[ht]
      \centering
      \includegraphics[width=2.2in]{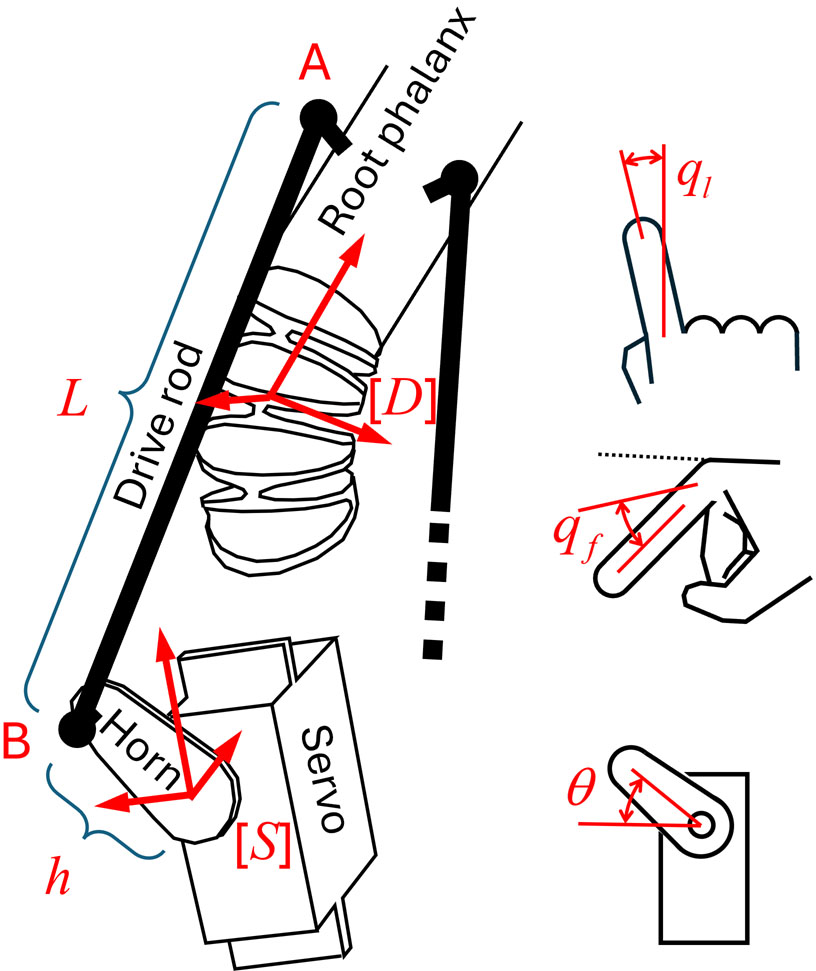}
      \caption{Notations for inverse kinematics of the root joint}
      \label{fig root joint IK}
\end{figure}

Figure~\ref{fig root joint IK} illustrates the notations.
Given the tilting angles of the root joint ${\bf{q}} \in \mathbb{R}^2$, the goal is to find the rotating angle $\theta$ of each servo.
Consider a digit frame $\left[D\right]$ attached to the root phalanx and a servo frame $\left[S\right]$ attached to the servo body.
Consider one of the drive rods and its two ends A and B.
%Suppose we have a digit frame $D$ attached to the root phalanx, a servo frame $S$ attached to the servo body, and a drive rod with ends A and B.
The position of point A in $\left[D\right]$ is a constant ${\bf{p}}_{_{A,D}} \in \mathbb{R}^3$.
Its representation in $\left[S\right]$ is 
\begin{equation}
{\bf{p}}_{_{A,S}} = {\bf{p}}_{_{A,D}}\cdot{\bf{R}}_{_{D2S}}\left( {\bf{q}} \right)+{\boldsymbol{\delta}}_{_{D2S}}
\end{equation}
where the rotation matrix ${\bf{R}}_{_{D2S}}$ is parameterized by $\bf{q}$, and the displacement vector $\boldsymbol{\delta}_{_{D2S}}$ is a constant (because $\left[D\right]$ is originated at the rotation center of the root joint).
Given~${\bf{q}}$, ${\bf{p}}_{_{A,S}} = \left[ x_{_A}, y_{_A}, z_{_A} \right]^T$ can be easily calculated using the coordinate transformation.
Meanwhile, the position of point~B in $\left[S\right]$ is ${\bf{p}}_{_{B,S}} = \left[ h\cos{\theta}, h\sin{\theta}, 0 \right]^T$, where $h$ is the length of the servo horn.
${\bf{p}}_{_{A,S}}$ and ${\bf{p}}_{_{B,S}}$ are related by the length of the drive rod 
\begin{equation}
    L = \left|\left| {\bf{p}}_{_{A,S}} - {\bf{p}}_{_{B,S}} \right|\right|_{_2}
\end{equation}
which gives
\begin{equation}
    \left(x_{_A} - h\cos{\theta}\right)^2 + \left(y_{_A} - h\sin{\theta}\right)^2 = L^2 - z_{_A}^2
\end{equation}
from where $\theta$ can be found using trigonometric addition identities.

The forward kinematics is much more complicated because the two drive rods cannot be considered individually like in the analysis of inverse kinematics.
Nevertheless, we have discovered that kinematics of the root joint can actually be well approximated by an affine mapping
\begin{equation}
    \bf{q} = 
    \begin{pmatrix}
        a & b \\
        c & d
    \end{pmatrix}
    \text{\large$\bm{\uptheta}$} +
    \begin{pmatrix}
        e \\
        f
    \end{pmatrix}
\end{equation}
%\[\left( {\begin{array}{*{20}{c}}
%a&b\\
%c&d
%\end{array}} \right)\]
where $\text{\large$\bm{\uptheta}$} \in \mathbb{R}^2$ includes rotation angles of the two servos, and $a,b,c,d,e,f$ are scalar coefficients.
We first use the inverse kinematics derived earlier to calculate the servo angles $\text{\large$\bm{\uptheta}$}$ of 10,000 samples of $\bf{q}$ across the motion range of the root joint.
The $\{{\bf{q}},\text{\large$\bm{\uptheta}$}\}$ pairs are then run through a least squares regression which identifies the coefficients as
$a = -0.2775$, $b = -0.2451$, $c = -0.2779$, $d = 0.2466$, $e = -6.902$, $f = 2.097$
with a very good regression R value at 0.981.
%$a = -0.3163$, $b = -0.2837$, $c = -0.3054$, $d = 0.2792$, $e = -9.6125$, $f = 3.0214$
%with a good regression R value at 0.916.

\begin{figure}[ht]
      \centering
      \includegraphics[width=3.1in]{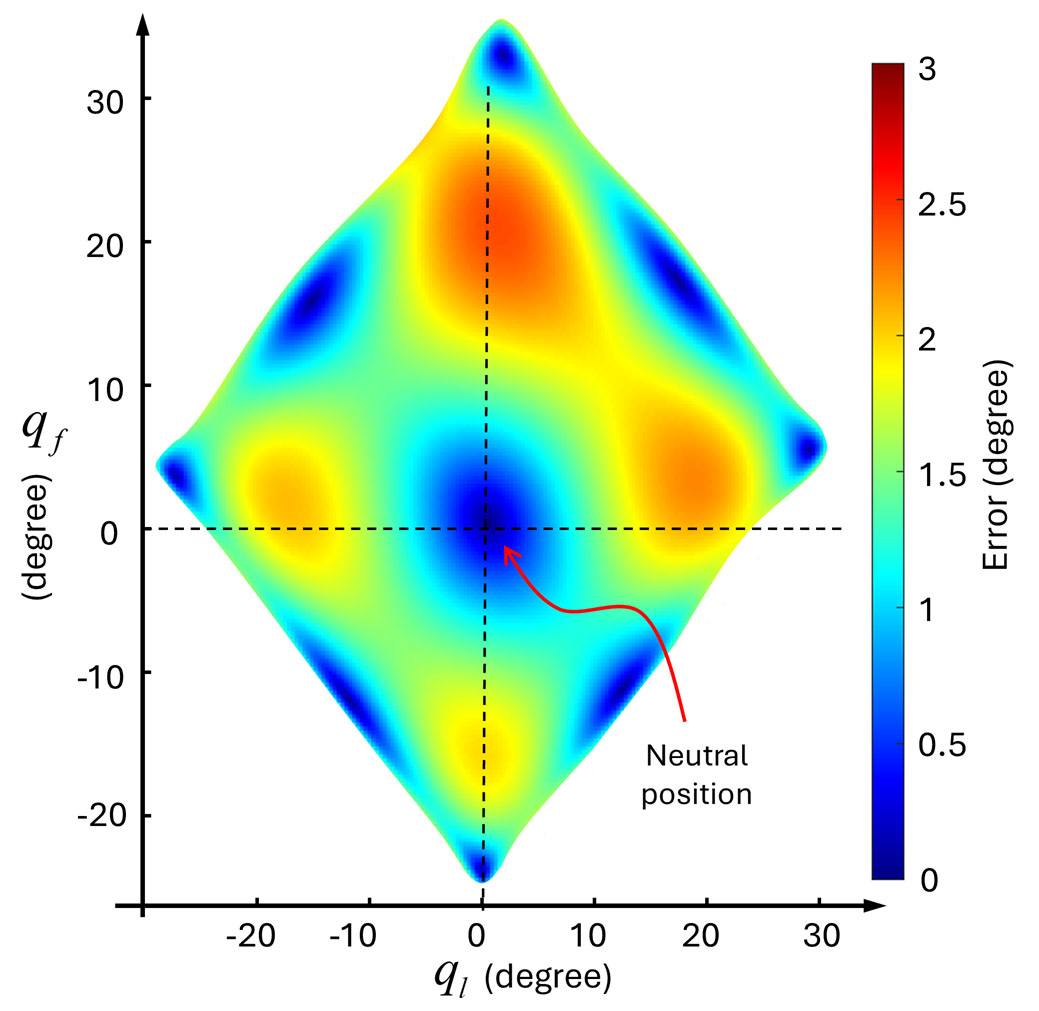}
      \caption{Error of an affine approximation of the root joint's kinematics. $q_f$ and $q_l$ are as defined in Fig.~\ref{fig root joint IK}}
      \label{fig approx IK err}
\end{figure}

The identified coefficients show a particular pattern: $a \approx b$ and $c \approx -d$, which suggests that flexion of the root joint is proportional to the average of the two servo angles while the lateral tilting is proportional to their difference.
This is valid because the servo horns are quite short and the drive rods spread out from the root phalanx by only small angles.
Note that there is no hard constraint enforcing $a=b$ and $c=-d$ in the regression because the two servos that drive the root joint are not exactly placed symmetrically and their drive rods are of different lengths.
Figure~\ref{fig approx IK err} shows the approximation errors across the motion range of the root joint (bounded by drivetrain singularity).
The error is calculated as $\left|\left|\hat{\bf{q}} - {\bf{q}}\right|\right|_{_2}$,
where $\hat{\bf{q}}$ is the estimate of ${\bf{q}}$ based on the affine approximation.
Majority of the motion range has an approximation error lower than $1.8^\circ$ while the largest error is less than $2.7^\circ$.

\begin{figure}[ht]
      \centering
      \includegraphics[width=3.3in]{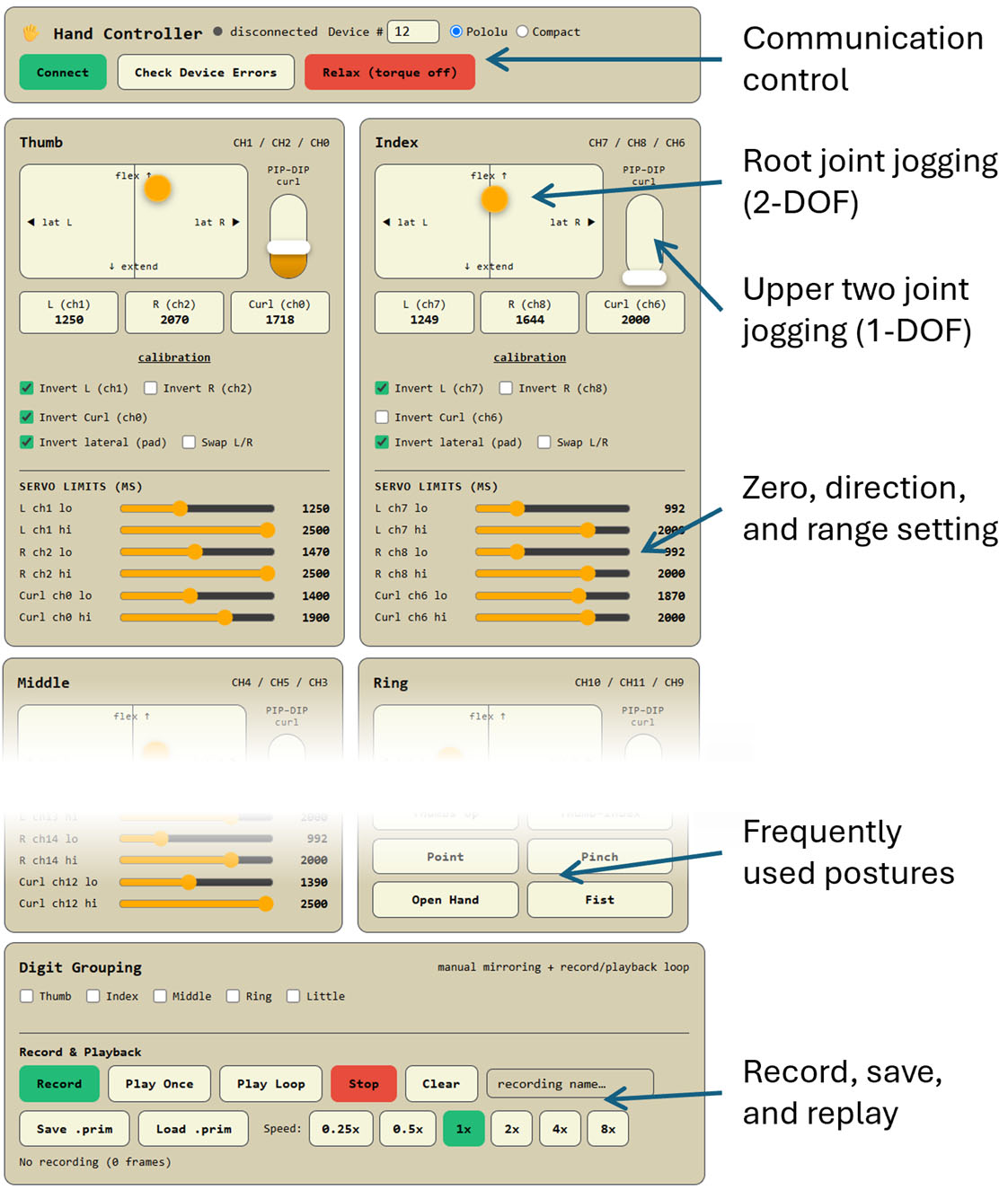}
      \caption{A GUI generated by AI}
      \label{fig gui screenshot}
      \vspace{-0.2cm}
\end{figure}

\section{AI-Aided Control}
\label{sec llm ctrl}
%The bottleneck after designing, performing the kinematic analysis, and fabricating the robot hand is the development of a control interface.
%The root joint of the finger is actuated collectively with two servos (with anchors attached in a V-shaped orientation, as shown in Fig. design parameters).
%This calls for a precise kinematic interface that can accept flexion and lateral values, and incorporate that on each finger.
%Additionally, there is also a need for motion primitives for performing different in-hand manipulation tasks such as reorientation, translation, twisting, etc.
Because of their high number of DOAs and nontrivial kinematics, controlling dexterous humanoid robot hands can be a technical hassle.
We have explored using the latest artificial intelligence (AI) to aid the development of control and find it surprisingly efficient.
%We address this by demonstrating that kinematic and hardware specifications can serve as a description/prompt to develop a control interface and motion primitives for the robot hand.
% Include prompt in appendix if allowed??
Claude (Sonnet 4.6 version) is used for its advanced coding and multi-modal reasoning ability.
%\subsection{GUI Development}
%Claude was used for the interface development, while our role was to mathematically derive the kinematics, design the mechanisms with dimensions, and perform physical verification, and physically refining the primitives. 
First, a graphical user interface (GUI) is developed.
The prompt consists of only plain text without any figures or CAD models, and includes the following specifications of the hand:
\begin{itemize}
    \item Lengths and thicknesses of the phalanges;
    \item Location and orientation of the digits on the palm;
    \item Closed-form equations of the inverse kinematics of the root joint as discussed in Section~\ref{sec kinematics};
    \item The coupling ratio of the upper two joints and the mapping between their flexion and the stroke of the elastic drive rod;
    \item Motion ranges of all DOAs;
    \item Make and model of the servo controller (Pololu / Mini Maestro 18) and the assignment of its channels to the servos;
\end{itemize}
as well as the desired GUI features:
\begin{itemize}
    \item For each digit, a 2-DOF pad for jogging the root joint, and a slider for jogging the flexion of the upper two joints;
    \item A calibration panel for setting the zeros, directions, and ranges of all DOAs;
    \item Buttons for recording hand postures and motion, save them to files, and playback.
\end{itemize}
%No additional specifications were provided other than the kinematics, hand topology, and panel design requirements.
With minimal manual adjustment, the GUI produced by Claude (Fig.~\ref{fig gui screenshot}) functions well with the physical prototype of the hand.

\begin{figure}[ht]
      \centering
      \includegraphics[width=2.7in]{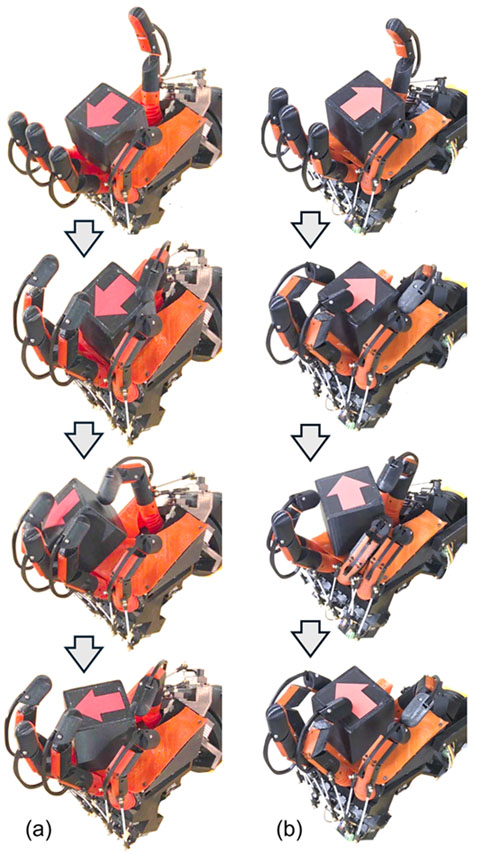}
      \caption{AI-generated motion trajectories - (a) draft, (b) revised}
      \label{fig AI generated trajectories}
\end{figure}

\begin{figure*}[ht]
      \centering
      \includegraphics[width=\textwidth]{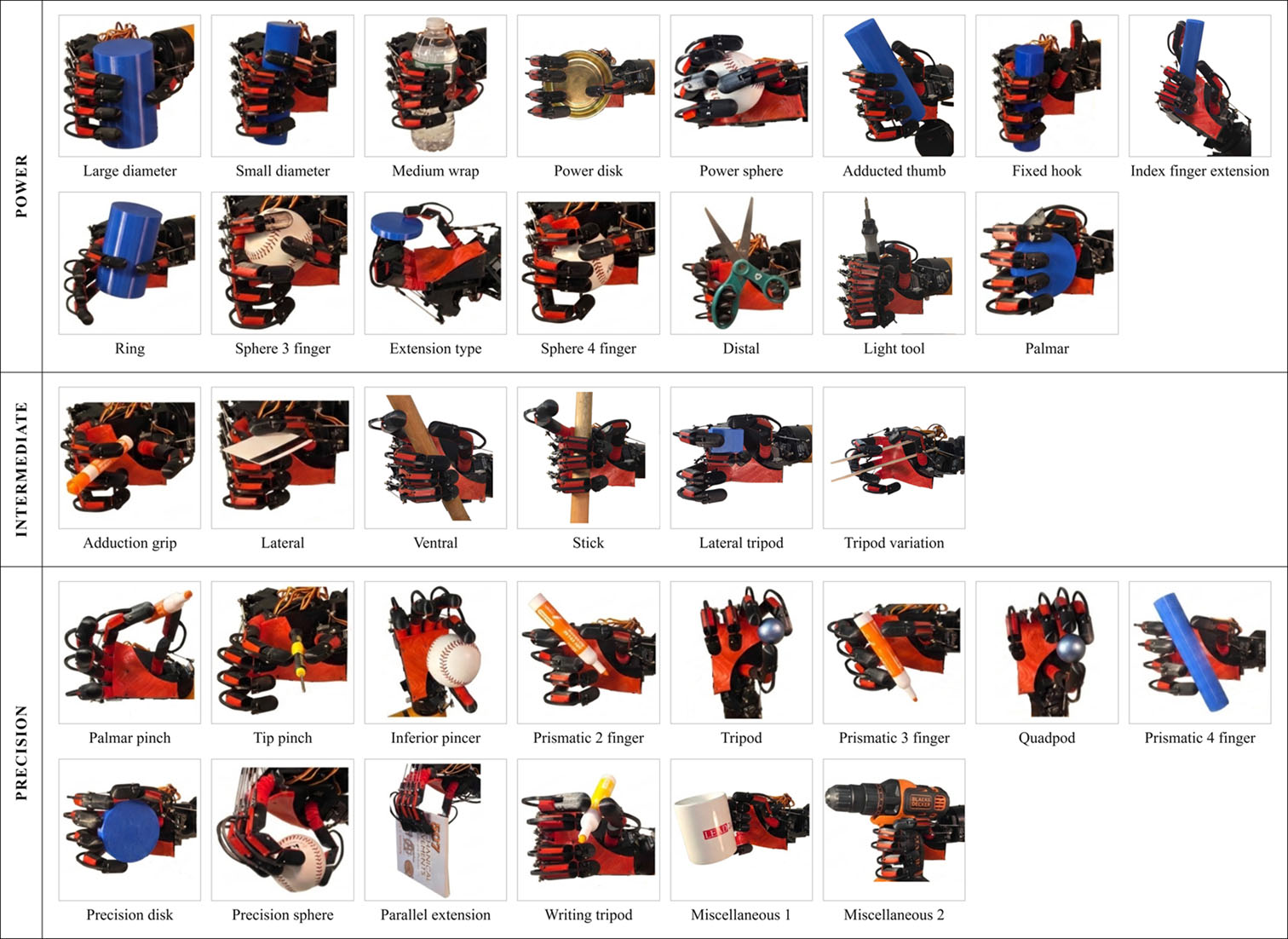}
      \caption{Validation of grasping ability according to the GRASP taxonomy \cite{feix2015grasp}}
      \label{fig grasping}
\end{figure*}

In addition, we have experimented using Claude to generate motion trajectories for advanced hand skills such as in-hand re-orientation of an object.
The prompt includes:
\begin{itemize}
    \item Description of the object - e.g., a cube of 55 mm edge length;
    \item Digits to be used actively (e.g., the thumb, index finger, ring finger) and digits to be kept static (e.g., middle finger extended, pinky finger curled);
    \item (Non-quantitative) description of the initial placement of the object on the palm - e.g., in the middle of the three active digits; and
    \item Description of the manipulation goal - e.g., spinning the cube with the palm facing up.
\end{itemize}

Without any additional information, Claude has produced a draft motion trajectory of 40 frames.
It does not function well because the digits cannot make proper contact with the object (Fig.~\ref{fig AI generated trajectories}-a).
The problem is described in a follow-up prompt to Claude for revision.
Claude responds by requesting the user to reset the initial placement of the object and use the GUI to jog the digits so that proper initial contacts are made.
The user-adjusted posture is then fed back to Claude, which revises the trajectory automatically.
The revised trajectory is surprisingly successful as shown in Fig.~\ref{fig AI generated trajectories}-b.

\section{Physical Validation}
\label{sec physical validation}
%We validate the proposed hand through a series of tests and experiments to assess its dexterity.
%We first evaluate thumb opposition across all digits, followed by 33 different grasp types listed in the Feix taxonomy, and then conclude with in-hand manipulation and tool-use demonstrations.
%These tests will verify that the proposed design achieves human-like dexterity across a wide range of grasping and manipulation tasks.

\subsection{Thumb Opposition}
\begin{figure}[thpb]
       \centering
       \vspace{0.5cm}
       \includegraphics[width=3.3in]{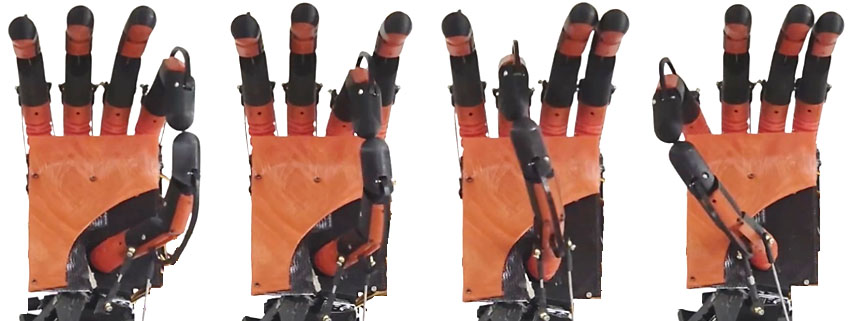}
       \caption{Thumb opposition}
       \label{fig thumb opposition}
%       \vspace{-0.4cm}
\end{figure}

Thumb opposition is the ability to move the thumb so that its tip touches the tips of the four fingers one by one.
Being able to do thumb opposition is considered a critical evolutionary advantage and a major milestone in early child development of fine motor skills.
It is of fundamental importance for realizing dexterous grasping and manipulation, and has been used commonly as a simple yet effective diagnosis of hand function deficits.
Figure~\ref{fig thumb opposition} shows thumb opposition performed by our prototype.

\subsection{Grasping}
The hand's grasping ability is assessed according to the GRASP taxonomy (a.k.a., the Feix taxonomy~\cite{feix2015grasp}).
It is widely recognized as the most comprehensive modern classification of human hand grasps, and has been commonly used in the evaluation of humanoid robot hands.
The taxonomy includes 33 grasp types and is organized by grasping strength (power, intermediate, precision) as well as opposition type, finger assignment, and thumb position.
A mix of real everyday objects and 3D printed mock objects are used in our tests. 
Figure~\ref{fig grasping} shows the grasps realized by our prototype.
Other than the 33 types in the taxonomy, we have extended our tests to grasping objects that have more particular features for functional grasps, such as a mug and a power drill.
Despite a good fulfillment in general, some of the grasps are not realized perfectly, mainly the Stick, Light tool, and Tripod variation (chopsticks).
The imperfection is caused by the limited flexion range of the fingers and lack of the thenar web space between the thumb and index finger.

\subsection{Object manipulation}
\begin{figure}[ht]
       \centering
       \includegraphics[width=3.4in, height=7in]{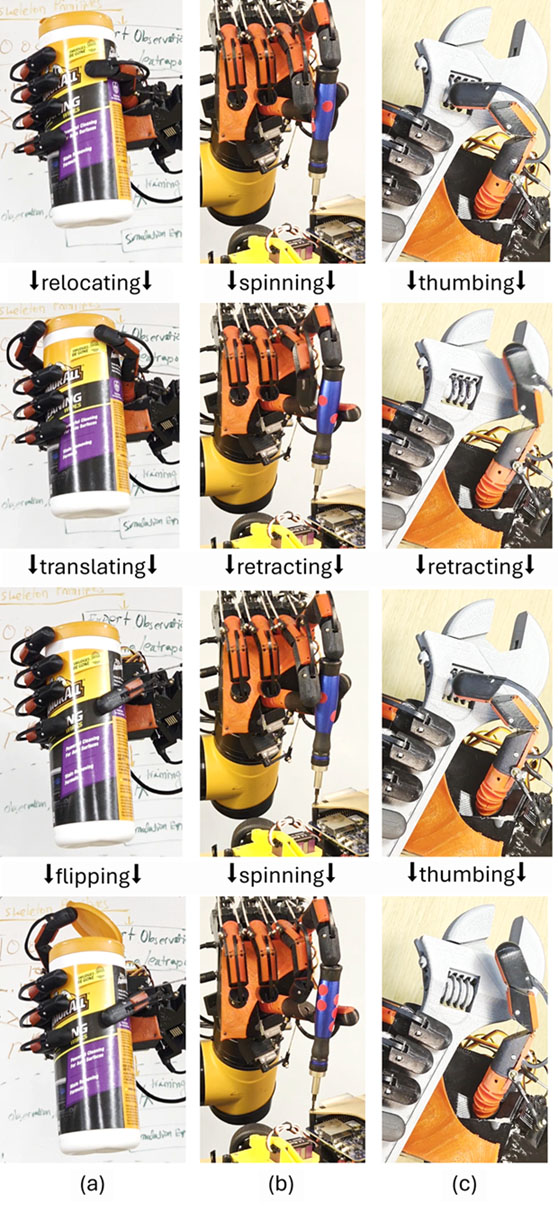}
       \caption{Object manipulation}
       \vspace{-0.4cm}
       \label{fig object manipulation}
\end{figure}

Other than in-hand object rotation shown earlier in Section~\ref{sec llm ctrl} Fig.~\ref{fig AI generated trajectories}, we present another three demonstrations of object manipulation using the hand.
In the first demonstration (Fig.~\ref{fig object manipulation}-a), the robot shifts an object in-hand so that its lid can be reached and flipped open by the index finger.
%An object of a larger size is used as a challenge.
The shifting validates that the thumb and index finger can coordinate with the other three digits in a humanoid way to reposition objects in-hand with fine precision.
The lid flipping challenges the motion range and strength of the index finger.
% This demonstrates its ability to perform multiple sub-tasks within a single sequence.
Figure~\ref{fig object manipulation}-b shows the hand using a precision screwdriver.
It requires the thumb and middle finger to coordinate delicately and alternate between retraction and spinning the screwdriver without pushing it out of the holding of the index finger.
Figure~\ref{fig object manipulation}-c shows the hand adjusting a crescent wrench by thumbing its worm screw.
In particular, the worm screw is thumbed outward to close the jaws.
This demonstrates effective bidirectional actuation of the digits.
The task cannot be accomplished with conventional tendon-driven designs where extension of the digits relies on spring-loaded auto-returning with limited strength\footnote{
Some advanced tendon-driven designs do have bidirectional actuation but require sophisticated cable routing systems or additional actuators.
This complication leads to issues on cost, reliability and durability.
}.
%There are more object manipulation tasks such as opening a lunch box, operating a keypad, and using a spray bottle, which are included in the supplemental video.
More demonstrations can be found in the supplementary video.

\section{Conclusions}
This paper presents a new design of linkage-driven dexterous humanoid robot hands.
Thanks to the use of a unique flexure hinge and elastic drive rods, the design requires much fewer parts than its peers and can be fabricated at a size closer to regular human hands with merely a budget 3D printer.
The advantage helps improve reliability, durability, cost-effectiveness, as well as lower the barrier to entry for using and developing dexterous humanoid robot hand functions.
In addition, the design greatly simplifies the kinematics of the drivetrains, which is beneficial to modeling, estimation, and control.
Dexterity of the hand is validated using the thumb opposition test, grasping according to Feix's GRASP taxonomy, and a variety of object manipulation tasks.
In addition, this paper presents our latest findings on kinematics analysis and AI-aided control.

As a preliminary reference for reliability and durability, at this point, we have conducted a total of around 200 hours of tests on the same prototype.
Majority of the tests ran over one hour, with the longest one lasted for two hours.
Throughout all the tests we have conducted, two mechanical glitches happened during the early stage of validation.
Both were caused by the breaking of an elastic drive rod.
One was because of a stock defect.
The other was because of improper tooling for shaping the connectors.
Other than those, no sign of mechanical degradation has been observed.

Our study will next address the following aspects:
\begin{itemize}
    \item Rigorous and quantitative study on reliability and durability, especially regarding fatigue of the elastic drive rods.
    \item Refining the topology of the hand to better resemble human hands.
    Other than adjusting the dimensions and motion ranges of the digits and their placement on the palm, this also includes reforming the palm surface and adding a thenar web space between the thumb and index finger.
    %The latter is crucial for improving the grasping.
    \item Optimizing the coupled kinematics of the upper two joints. In particular, the coupling rod shall be redesigned to go through the middle phalanx instead of from outside.
    Its shaping shall be studied in detail to obtain a proper nonlinear flexibility for passive compliance.
    \item Redesigning the root joint of the thumb. Specially, the 2-DOF flexure hinge shall be redesigned to couple an axial rotation to the lateral swing of the thumb.
    \item Exploring 3D printing settings for reliably printing a whole digit body in one piece while achieving proper flexibility at the joints and stiffness at the phalanges.
    This will further simplify fabrication and ready the design for open sourcing.
    \item Leveraging the backdrivability of the drivetrains to explore the possibility of contact estimation based on servo readings without using any touch sensors.
\end{itemize}

\vspace{0.2cm}

%\section*{APPENDIX}

%\section*{ACKNOWLEDGMENT}
%XXXXX XXXXX XXXXXX

%%%%%%%%%%%%%%%%%%%%%%%%%%%%%%%%%%%%%%%%%%%%%%%%%%%%%%%%%%%%%%%%%%%%%%%%%%%%%%%%

\bibliography{2_references}

@inproceedings{adwani2026dexterous,
  title={Dexterous Robot Hands with Linkage-Driven Three-{DOF} Fingers},
  author={Adwani, Neel and Ouali, Saffia and Orlowski, Mateusz and Liu, Ethan and Wang, Cong},
  booktitle={The 6th Modeling, Estimation and Control Conference},
  year={2026}
}

@inproceedings{kuo2010contribution,
  title={Contribution of passive properties of muscle-tendon units to the metacarpophalangeal joint torque of the index finger},
  author={Kuo, Pei-Hsin and Deshpande, Ashish D},
  booktitle={the 3rd IEEE RAS \& EMBS International Conference on Biomedical Robotics and Biomechatronics},
  pages={288--294},
  year={2010}
}

@inproceedings{ma2014underactuated,
  title={An underactuated hand for efficient finger-gaiting-based dexterous manipulation},
  author={Ma, Raymond R and Dollar, Aaron M},
  booktitle={IEEE International Conference on Robotics and Biomimetics},
  pages={2214--2219},
  year={2014}
}

@inproceedings{romero2024eyesight,
  title={Eyesight hand: {D}esign of a fully-actuated dexterous robot hand with integrated vision-based tactile sensors and compliant actuation},
  author={Romero, Branden and Fang, Hao-Shu and Agrawal, Pulkit and Adelson, Edward},
  booktitle={IEEE/RSJ International Conference on Intelligent Robots and Systems},
  pages={1853--1860},
  year={2024}
}

@article{hernando2023mechatronic,
  title={Mechatronic design of a self-contained dexterous robotic hand for gestural communication},
  author={Hernando, Miguel and Morillo, Carlos and Guffanti, Diego and Brunete, Alberto},
  journal={International Journal of Social Robotics},
  volume={15},
  number={2},
  pages={221--231},
  year={2023},
  publisher={Springer}
}

@article{shaw2023leap,
  title={{LEAP} hand: {L}ow-cost, efficient, and anthropomorphic hand for robot learning},
  author={Shaw, Kenneth and Agarwal, Ananye and Pathak, Deepak},
  journal={arXiv:2309.06440},
  year={2023}
}

@article{piazza2019century,
  title={A century of robotic hands},
  author={Piazza, Cristina and Grioli, Giorgio and Catalano, Manuel G and Bicchi, Antonio},
  journal={Annual Review of Control, Robotics, and Autonomous Systems},
  volume={2},
  number={1},
  pages={1--32},
  year={2019},
  publisher={Annual Reviews}
}

@article{teslahand,
  title         = "Tesla's humanoid robot delayed again because its hands can only be used for six weeks",
  author        = "TechWire",
  journal       = "TechWire",
  url           = "https://www.asiatechwire.com/Teslas-humanoid-robot-delayed-again-because-its-hands-can-only-be-used-for-six-weeks.html",
  month         = oct,
  year          = "2025"
}

@article{hong2018kulex,
  title={{KULEX}-hand: {A}n underactuated wearable hand for grasping power assistance},
  author={Hong, Man Bok and Kim, Sin Jung and Ihn, Yong Seok and Jeong, Gu-Cheol and Kim, Keehoon},
  journal={IEEE Transactions on Robotics},
  volume={35},
  number={2},
  pages={420--432},
  year={2018}
}

@article{liu2017development,
  title={Development of a humanoid robot hand with coupling four-bar linkage},
  author={Liu, Xinhua and Zheng, Xianhua and Li, Shengpeng},
  journal={Advances in Mechanical Engineering},
  volume={9},
  number={1},
  pages={1687814016686313},
  year={2017},
  publisher={SAGE Publications}
}

@article{yoon2017underactuated,
  title={Underactuated finger mechanism using contractible slider-cranks and stackable four-bar linkages},
  author={Yoon, Dukchan and Choi, Youngjin},
  journal={IEEE/ASME Transactions on Mechatronics},
  volume={22},
  number={5},
  pages={2046--2057},
  year={2017}
}

@article{cerulo2017teleoperation,
  title={Teleoperation of the {SCHUNK} {S5FH} under-actuated anthropomorphic hand using human hand motion tracking},
  author={Cerulo, Ilaria and Ficuciello, Fanny and Lippiello, Vincenzo and Siciliano, Bruno},
  journal={Robotics and Autonomous Systems},
  volume={89},
  pages={75--84},
  year={2017},
  publisher={Elsevier}
}

@article{wen2025dexterous,
  title={Dexterous Teleoperation of 20-{DoF} {ByteDexter} Hand via Human Motion Retargeting},
  author={Wen, Ruoshi and Zhang, Jiajun and Chen, Guangzeng and Cui, Zhongren and Du, Min and Gou, Yang and Han, Zhigang and Hu, Junkai and Huang, Liqun and Niu, Hao and others},
  journal={arXiv:2507.03227},
  year={2025}
}

@article{kashef2020robotic,
  title={Robotic hand: {A} review on linkage-driven finger mechanisms of prosthetic hands and evaluation of the performance criteria},
  author={Kashef, S Reza and Amini, Samane and Akbarzadeh, Alireza},
  journal={Mechanism and Machine Theory},
  volume={145},
  pages={103677},
  year={2020},
  publisher={Elsevier}
}

@inproceedings{ueda2005development,
  title={Development of the {NAIST-Hand} with vision-based tactile fingertip sensor},
  author={Ueda, Jun and Ishida, Yutaka and Kondo, Masahiro and Ogasawara, Tsukasa},
  booktitle={IEEE International Conference on Robotics and Automation},
  pages={2332--2337},
  year={2005}
}

@article{kim2021integrated,
  title={Integrated linkage-driven dexterous anthropomorphic robotic hand},
  author={Kim, Uikyum and Jung, Dawoon and Jeong, Heeyoen and Park, Jongwoo and Jung, Hyun-Mok and Cheong, Joono and Choi, Hyouk Ryeol and Do, Hyunmin and Park, Chanhun},
  journal={Nature Communications},
  volume={12},
  number={1},
  pages={7177},
  year={2021}
}

@inproceedings{mouri2002anthropomorphic,
  title={Anthropomorphic robot hand: {G}ifu hand {III}},
  author={Mouri, Tetsuya and Kawasaki, Haruhisa and Yoshikawa, Keisuke and Takai, Jun and Ito, Satoshi and others},
  booktitle={International Conference on Control Automation and Systems},
  pages={1288--1293},
  year={2002}
}

@article{bhatt,
  author = {Revant N. Bhatt},
  title = {Quantifying Serviceability: {A} Maintainability Framework for Tendon-Driven Robotic Hands},
  journal = {International Journal of Science and Research},
  volume = {14},
  number = {9},
  year = {2025},
  month = sep,
  pages = {1527--1530}
}

@article{jeong,
author = {Useok Jeong and Keunsu Kim and Sang-Hun Kim and Hyunhee Choi and Byeng Dong Youn and Kyu-Jin Cho},
title ={Reliability analysis of a tendon-driven actuation for soft robots},
journal = {The International Journal of Robotics Research},
volume = {40},
number = {1},
pages = {494-511},
year = {2021}
}

@article{wangandli,
author = {Yiwei Wang  and Wenyang Li  and Shunta Togo  and Hiroshi Yokoi  and Yinlai Jiang },
title = {Survey on Main Drive Methods Used in Humanoid Robotic Upper Limbs},
journal = {Cyborg and Bionic Systems},
volume = {2021},
number = {},
pages = {},
year = {2021}
}

@article{iitsofthand,
author = {Catalano, M.G. and Grioli, G. and Farnioli, E. and Serio, A. and Piazza, C. and Bicchi, A.},
title = {Adaptive synergies for the design and control of the {Pisa/IIT} {SoftHand}},
year = {2014},
publisher = {Sage Publications, Inc.},
address = {USA},
volume = {33},
number = {5},
issn = {0278-3649},
journal = {International Journal of Robotics Research},
month = apr,
pages = {768–782},
numpages = {15}
}

@article{feix2015grasp,
  title={The {GRASP} taxonomy of human grasp types},
  author={Feix, Thomas and Romero, Javier and Schmiedmayer, Heinz-Bodo and Dollar, Aaron M and Kragic, Danica},
  journal={IEEE Transactions on human-machine systems},
  volume={46},
  number={1},
  pages={66--77},
  year={2015}
}

@article{wen2025gr,
  title={{GR}-{D}exter Technical Report},
  author={Wen, Ruoshi and Chen, Guangzeng and Cui, Zhongren and Du, Min and Gou, Yang and Han, Zhigang and Huang, Liqun and Lei, Mingyu and Li, Yunfei and Li, Zhuohang and others},
  journal={arXiv:2512.24210},
  year={2025}
}

@article{roda2022studying,
  title={Studying kinematic linkage of finger joints: {E}stimation of kinematics of distal interphalangeal joints during manipulation},
  author={Roda-Sales, A. and Sancho-Bru, J. L. and Vergara, M.},
  journal={PeerJ},
  volume={10},
  pages={e14051},
  year={2022}
}

@inproceedings{liu2022three,
  title={A three-finger dexterous hand with suction cups integrated on fingertips},
  author={Liu, Yuanbo and Lu, Wenjie and Chen, Diansheng and Guo, Wei Dong},
  booktitle={the 12th International Conference on CYBER Technology in Automation, Control, and Intelligent Systems},
  pages={983--988},
  year={2022},
  organization={IEEE}
}

@article{zaw2025design,
  title={Design and kinematic analysis of a novel series-parallel hybrid finger for robotic hands},
  author={Zaw, Thiha and Mahalingam, Dasharadhan and Baiata, Nicholas and Patankar, Aditya and Chakraborty, Nilanjan},
  journal={Journal of Mechanisms and Robotics},
  volume={17},
  number={4},
  pages={044512},
  year={2025},
  publisher={ASME}
}
\bibliographystyle{IEEEtran}

\end{document}